\documentclass[conference]{IEEEtran}
\IEEEoverridecommandlockouts

\makeatletter
\def\ps@IEEEtitlepagestyle{%
  \def\@oddfoot{\mycopyrightnotice}%
  \def\@oddhead{\hbox{}\@IEEEheaderstyle\leftmark\hfil\thepage}\relax
  \def\@evenhead{\@IEEEheaderstyle\thepage\hfil\leftmark\hbox{}}\relax
  \def\@evenfoot{}%
}
\def\mycopyrightnotice{%
  \begin{minipage}{\textwidth}
  \centering \scriptsize
  Copyright~\copyright~2023 IEEE. Personal use of this material is permitted. Permission from IEEE must be obtained for all other uses, in any current or future media, including\\reprinting/republishing this material for advertising or promotional purposes, creating new collective works, for resale or redistribution to servers or lists, or reuse of any copyrighted component of this work in other works by sending a request to pubs-permissions@ieee.org.
  \end{minipage}
}
\makeatother

\usepackage{graphicx}
\usepackage{textcomp}
\usepackage{xcolor}
\usepackage{amsmath,amssymb}
\usepackage{makecell}
\usepackage{multirow}
\usepackage{caption}
\usepackage{subcaption}
\usepackage{float}
\usepackage{graphicx}
\usepackage{capt-of}% or \usepackage{caption}
\usepackage{booktabs}
\usepackage{varwidth}
\usepackage{amsmath}
\usepackage{algorithm,algorithmic}
\usepackage{setspace, etoolbox}
\usepackage{hyperref}
\usepackage{subcaption}

\usepackage{cite}
\usepackage{amsmath,amssymb,amsfonts}
\usepackage{algorithmic}
\usepackage{graphicx}
\usepackage{textcomp}
\usepackage{xcolor}
\usepackage{makecell}
\usepackage{booktabs}
\newcommand{\citet}[1]{\citeauthor{#1}~(\citeyear{#1})}
\newcommand{\citep}[1]{\cite{#1}}

\def\BibTeX{{\rm B\kern-.05em{\sc i\kern-.025em b}\kern-.08em
    T\kern-.1667em\lower.7ex\hbox{E}\kern-.125emX}}

\makeatletter
\newcommand{\linebreakand}{%
  \end{@IEEEauthorhalign}
  \hfill\mbox{}\par
  \mbox{}\hfill\begin{@IEEEauthorhalign}
}
\makeatother

\newcommand\T{\mathcal{T}}

\newcommand\M{\mathcal{M}}

\newcommand\SN{\mathcal{S}}
\newcommand\V{\mathcal{V}}

\newcommand\D{\mathcal{D}}

\newcommand\DS{\mathcal{DS}}
\DeclareMathOperator*{\argmax}{arg\,max}

\begin{document}

\title{Deep Sensitivity Analysis for Objective-Oriented Combinatorial Optimization
%{\footnotesize \textsuperscript{*}Note: Sub-titles are not captured in Xplore and should not be used}
}

\author{\IEEEauthorblockN{1\textsuperscript{st} Ganga Gireesan}
\IEEEauthorblockA{\textit{Mississippi State University} \\
\textit{Mississippi State, USA}\\
gg733@msstate.edu}
\and
\IEEEauthorblockN{1\textsuperscript{st} Nisha Pillai}
\IEEEauthorblockA{\textit{Mississippi State University} \\
\textit{Mississippi State, USA}\\
pillai@cse.msstate.edu}
\and
\IEEEauthorblockN{2\textsuperscript{nd} Michael J Rothrock}
\IEEEauthorblockA{\textit{USDA-ARS} \\
\textit{Athens, GA, USA} \\
michael.rothrock@usda.gov} 
\linebreakand
\IEEEauthorblockN{3\textsuperscript{rd} Bindu Nanduri}
\IEEEauthorblockA{\textit{Mississippi State University} \\
\textit{Mississippi State, USA}\\
nanduribindu@gmail.com}

\and
\IEEEauthorblockN{4\textsuperscript{th} Zhiqian Chen}
\IEEEauthorblockA{\textit{Mississippi State University} \\
\textit{Mississippi State, USA}\\
zchen@cse.msstate.edu}
\and
\IEEEauthorblockN{5\textsuperscript{th} Mahalingam Ramkumar}
\IEEEauthorblockA{\textit{Mississippi State University} \\
\textit{Mississippi State, USA}\\
ramkumar@cse.msstate.edu}
}
\maketitle

\IEEEpubidadjcol

\begin{abstract}
Pathogen control is a critical aspect of modern poultry farming, providing important benefits for both public health and productivity. Effective poultry management measures to reduce pathogen levels in poultry flocks promote food safety by lowering risks of food-borne illnesses. They also support animal health and welfare by preventing infectious diseases that can rapidly spread and impact flock growth, egg production, and overall health. This study frames the search for optimal management practices that minimize the presence of multiple pathogens as a combinatorial optimization problem. Specifically, we model the various possible combinations of management settings as a solution space that can be efficiently explored to identify configurations that optimally reduce pathogen levels. This design incorporates a neural network feedback-based method that combines feature explanations with global sensitivity analysis to ensure combinatorial optimization in multiobjective settings. Our preliminary experiments have promising results when applied to two real-world agricultural datasets. While further validation is still needed, these early experimental findings demonstrate the potential of the model to derive targeted feature interactions that adaptively optimize pathogen control under varying real-world constraints.
\end{abstract}


\section{Introduction}

Combinatorial optimization (CO) tackles discrete decision and planning challenges across many fields, including route planning, scheduling, and resource allocation. The search spaces for these NP-hard problems are vast and exponential. Researchers employ a multifaceted approach to address the complexity. A number of algorithms are employed in this field, including exact algorithms~\cite{xu2020deep, Pillai_CO2023} that ensure optimality, but are slow, as well as approximate algorithms~\cite{weissteiner2023bayesian, acampora2023genetic} that rapidly find near-optimal solutions. Using a combination of mathematical, algorithmic, and artificial intelligence approaches, high-quality solutions can be found to complex combinatorial optimization problems in areas such as navigation, planning, and feature optimization.

In this research, we propose a novel approach to combinatorial optimization that incorporates associative feature importance weights, enabling optimization towards specific goals. As part of our method, explainable artificial intelligence techniques are incorporated with combinatorial optimization techniques to achieve multiple objectives. This provides greater flexibility compared to existing traveling salesman problem~\cite{yang2023memory} formulations.

Our key contribution is Deep Sensitivity Analysis (DS), a neural network based explainable AI~\cite{gunning2019xai} methodology for combinatorial optimization. It utilizes variance-based sensitivity analysis to train a neural network to optimize feature combination selection. Comparisons show improved performance over dynamic programming on two poultry food-borne pathogen datasets. This underscores the importance of incorporating explainable knowledge into the field of combinatorial optimization when addressing complex real-world challenges.

The outline of this paper is as follows. Related research is briefed under section~\ref{related}. Proposed approach is presented in section~\ref{approach}. Section~\ref{dataset} examines two multi-objective datasets used in this research to evaluate the problem. The experiments and results are described in section~\ref{experiments}. Conclusions are offered in section~\ref{conclusion}.

\begin{figure*}[t]
\centering
\includegraphics[width=0.95\textwidth]{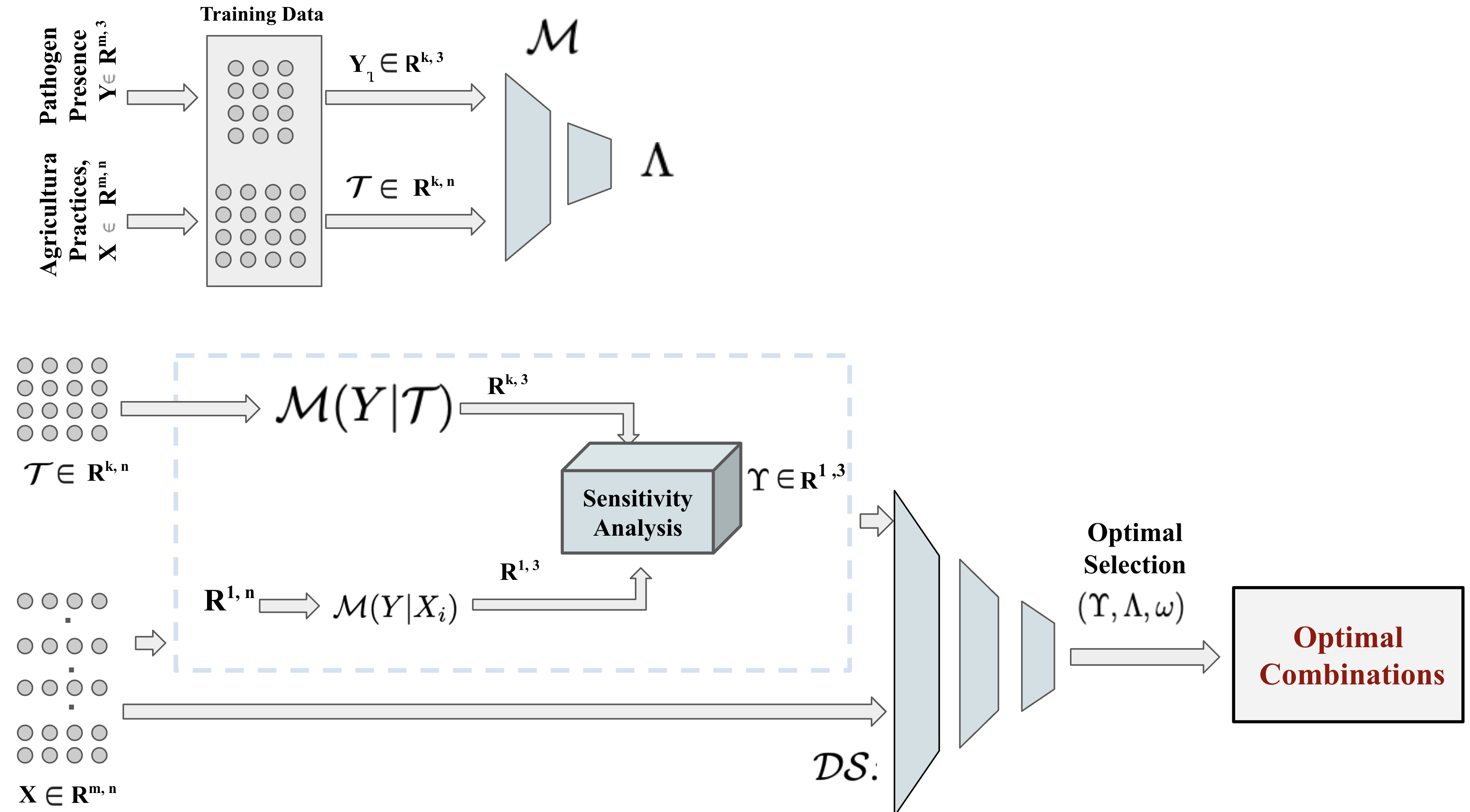}
\caption{Design diagram of Deep Sensitivity Analysis multi-label combinatorial optimization (DS). To predict food-borne pathogens (\textit{Salmonella}, \textit{Listeria}, and \textit{Campylobacter}) in the first stage of the process, we build a multi-label multi-layer perceptron architecture ($\M$) based on  farm practices. In our initial dataset, $m$ samples and $n$ features are represented as $X$ and $m$ samples and 3 pathogens are represented as $Y$. Samples are divided into $k$ train samples and $m - k$ test samples.  As a next step, we will construct an explainable artificial intelligence (AI) based Deep Sensitivity Analysis (DS) neural network. We use the training sample ($\T$) as a reference dataset, and used the model $\M$ to predict the pathogen content. Similarly, the pathogen content for every sample in the original dataset is predicted using the $\M$ model. These two predictions are used to calculate the global sensitivity score, which is then used to build the Deep Sensitivity Analysis (DS) neural network, together with agricultural samples. In the final step of our study, we learn the global explanation of feature combinations through objective-based selection.}

%COMMENTS: You dont have to go in to minute details like replication in the figure (you can do it elsewhere if you choose. Remove the duplication block - the duplication can be seen as inside the sensitivity analysis block. You can describe that if ou choose elsewhere.)
%Nisha -> I have explained about this replication in section III-C, 1st paragraph 2nd last sentence. 

%The pathogen content of each sample in the original dataset is predicted using model $\M$, and then this single prediction is repeated to match the dimensions of the training sample prediction.
%Design diagram of Deep Sensitivity Analysis multi-label combinatorial optimization. To predict food-borne pathogens (salmonella, listeria and campylobacter) in the first stage of the process, we build a multi-label multi-layer perceptron architecture based on our agricultural practices. To facilitate the implementation of explainable AI solutions in combinatorial optimization tasks, we propose a deep neural network system with global sensitivity analysis as a learning function. The training sample (T) serves as a reference dataset for comparing the feature to when determining the variance of the predictions. In the final step of our study, we learn the global explanation of feature combinations through objective-based selection. 

\label{fig:design}
\end{figure*}

\section{Related Research} \label{related}

The nature of combinatorial optimization problems makes them intrinsically challenging in real-life scenarios~\cite{kaya2022review}. Researchers over the years have made considerable efforts to address diverse problem classes~\cite{weinand2022research}, such as the quadratic assignment problem~\cite{hurtado2022exact}, the minimum spanning tree problem~\cite{majumder2022multi}, the location-routing problem~\cite{tahami2022literature}, and the traveling salesman problem~\cite{panwar2023discrete}, using both exact and approximate solutions. Integer programming~\cite{zhang2023survey} and branch-and-bound~\cite{zhang2022deep} have traditionally been used to solve CO problems, but the NP-hard nature of many combinatorial problems often makes exact solutions impractical. In these cases, metaheuristic algorithms, such as evolutionary algorithms~\cite{de2022random}, are valuable tools for solving NP-hard problems~\cite{zhao2022evolution} in computational optimization. Feedback-driven algorithms such as physics-based algorithms~\cite{geist2023combinatorial}, swarm-based algorithms~\cite{zhu2022improved}, bio-inspired algorithms, and nature-inspired algorithms apply feedback to find suboptimal solutions. In this research, deep learning is used to learn feature combination explanations to derive sub-optimal solutions to multi-objective problems.

Explainable AI techniques~\cite{laato2022explain} provide model transparency through simplification, feature relevance, local explanations, and visualizations. Leveraging explainability~\cite{panigutti2022understanding, ahmed2022artificial} allows optimization aligned with domain objectives. We employ variance-based global sensitivity analysis for feature selection based on relevance to specified goals. Research shows sensitivity analysis effectively explains nonlinear model behavior in areas like image processing and agriculture. The sensitivity-driven optimization provides interpretable solutions tailored to target outcomes.

%Generally, explainable AI algorithms can be classified into four main groups: 1) explanations by simplification, 2) explanations by feature relevance, 3) local explanations, and 4) explanations by visuality. Transparent and explainable methods can serve a wide range of purposes, as demonstrated by our work on using feature relevance explanations in agricultural systems to reduce the presence of food-borne pathogens. A variance-based global sensitivity analysis is used in this study to distinguish the reference dataset from the testing sample. 

Variance analysis has been successfully used by researchers in a number of different disciplines to measure both uncertainty~\cite{dahabreh2023sensitivity} and confidence~\cite{zhao2019sensitivity}. It is a proven method for analyzing parameter values in a transparent and interpretable manner~\cite{huang2022variance}. As an example, sensitivity analysis based on neural networks has been demonstrated to be effective in modeling plant processes~\cite{aouichaoui2022deepgsa}. As recommended by~\cite{shreim2023trainable}, Sobol sensitivity analysis fits well in image processing and rotation error calculations~\cite{jin2023sensitivity}. In general, variance-based sensitivity analysis is effective in addressing problems that require explanation through artificial intelligence.

\begin{figure*}[t]
\centering
\includegraphics[width=0.95\textwidth]{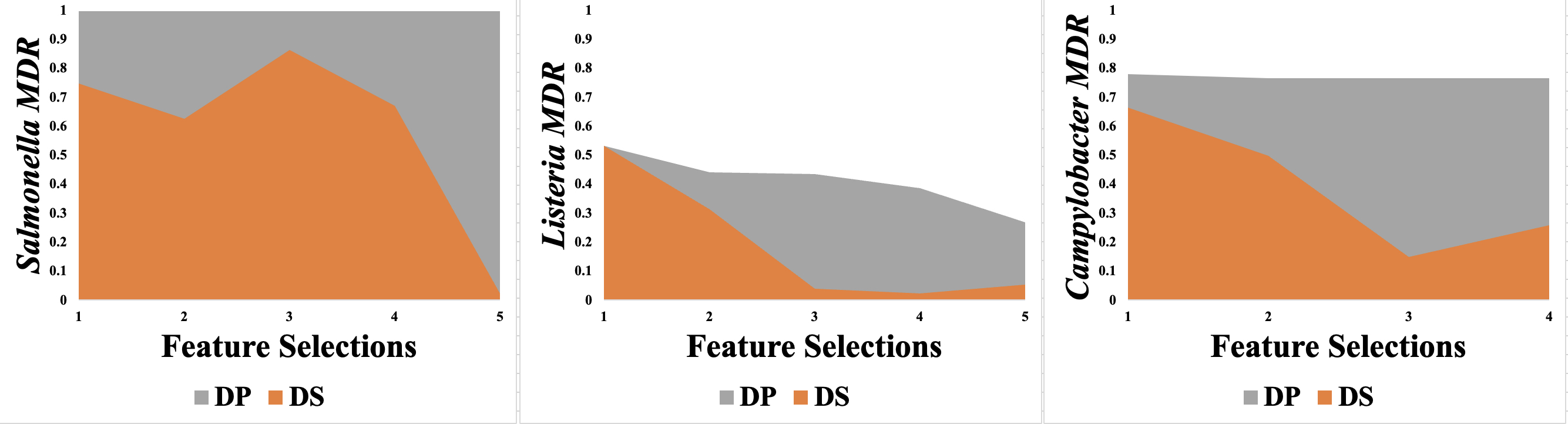}
\caption{Reduction of MDR based on pre-harvest agricultural practices. The results indicated that highly unbalanced, complex datasets required extensive analysis to interpret their feature associations to achieve optimization in the initial stages. However, when compared to dynamic programming (DP), DS significantly reduces \textit{Salmonella}, \textit{Listeria}, and \textit{Campylobacter} MDR at earlier stages of poultry production.}
\label{fig:results_mdr}
\end{figure*}

\begin{figure*}[h]
\centering
\includegraphics[width=0.95\textwidth]{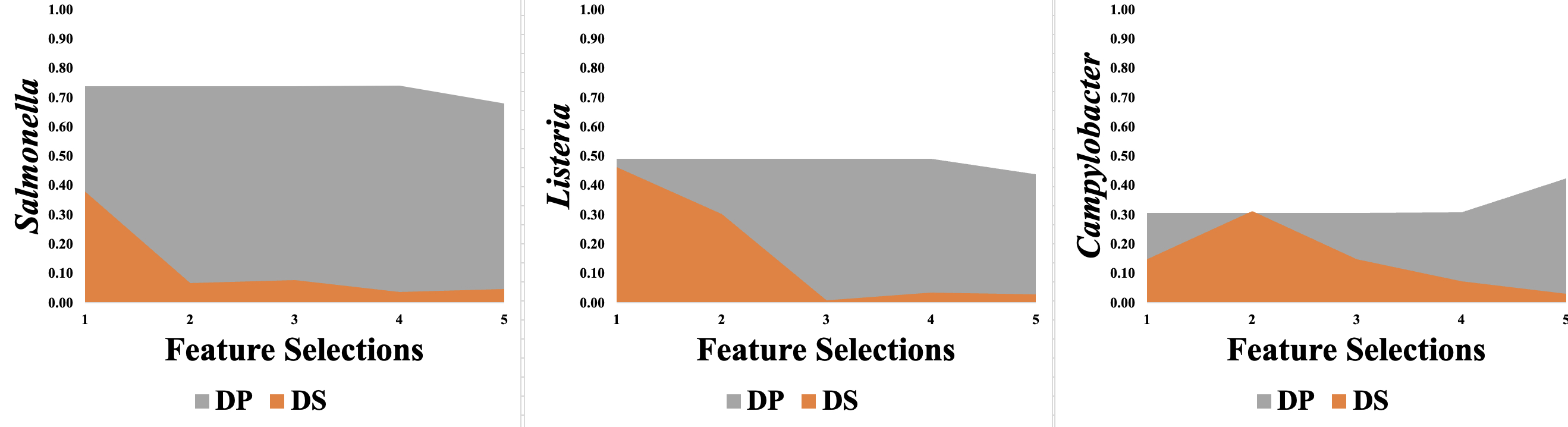}
\caption{Reduction of pathogen presence based on post-harvest agricultural practices. Results indicate that compared to dynamic programming (DP) algorithms, our proposed approach provides promising results in combinatorial optimization. Moreover, it shows that explainable AI that provides feature combination explanations can improve the decision-making during combinatorial optimization problems. }
\label{fig:results_pathogen}
\end{figure*}

\section{Approach}
\label{approach}

This study proposes a neural network based sensitivity analysis framework for determining optimal feature value combinations in agricultural settings (See Fig.~\ref{fig:design})). There are three parts to the framework: the first builds a multi-label classification network to predict the pathogen presence. Next, a variance-based global sensitivity score is calculated, while the third part creates a regression neural network to calculate global sensitivity. Our objective is determined at the testing stage based on the sensitivity score.

\paragraph{Variables}  This research uses dataset $\D$, which consists of agricultural data as input $X \in R^{ m, n}$ and pathogen presence as output variables $Y \in R^{ m, 3}$. Here, $m$ is the sample size and $n$ is the feature size. In this problem, we are interested in finding the optimal combination of features ($X$) and values ($\V$)  that will meet a specific objective. The dataset consisting of $<X$, $Y>$ is initially divided into train and test samples. The train dataset $\T$ is then used as the reference dataset to calculate the global sensitivity score.

\paragraph{Neural Network, $\M$} We use a single hidden layer neural network to simplify the architecture to predict $Y$ based on input variables $X$. The last layer of the classification algorithm uses sigmoid nonlinear activation, and a binary
cross-entropy loss is used for backpropagation to determine the prediction probabilities, $\Lambda$. This study utilizes our trained model $\M$ to determine the importance of features and values.

\paragraph{Deep Global Sensitivity Analysis, $\DS$}  The key contribution of our work is the use of deep neural networks to calculate the global sensitivity score. The deep sensitivity analysis network ($\DS$) consists of two or more hidden layers and we expect to generate a variation-based sensitivity score, $\Upsilon_i$, as a regression output for an input sample, $X_i$. Our combinatorial optimization model is based on feature importance. If any variation in input results in a substantial variation in output, the sensitivity score will be high, indicating that the input has a high importance in the model prediction. Also, the uncertainty associated with $Y$ can be attributed to the uncertainty associated with $X_i$ since it represents most of its variance. A variance-based global sensitivity analysis~\cite{azzini2021sobol} is employed in this research to determine the influence of feature associations in predictions. While training, we repeat
a single sample $X_i$ to match the dimensions of the reference train set $\T$ to calculate sensitivity score. In the neural network learning process, variation from the reference train set $\T$ is used to determine the loss function.

\begin{equation}
\Upsilon \gets \dfrac{COV(\M_i, \M_T)}{VAR(\M_T)}
\end{equation}

where $\M_i$ is $\M(Y|X_i)$ and $\M_T$ is $\M(Y|\T)$.

During testing, we need to learn the importance of each combination of features ($X_{q1}, X_{q2}.....$), so we clone the reference test set $\T$ and modify it to fix respective features ($X_{q1}, X_{q2}.....$) with its value in $X$ and use this input sample $X_c$ to predict using $\DS$ how important each feature is.

\begin{equation}
X_c \gets T_1......T_k, X_{q1}, X_{q2}.....T_n
\end{equation}

\begin{figure*}[h]
\centering
\includegraphics[width=0.95\textwidth]{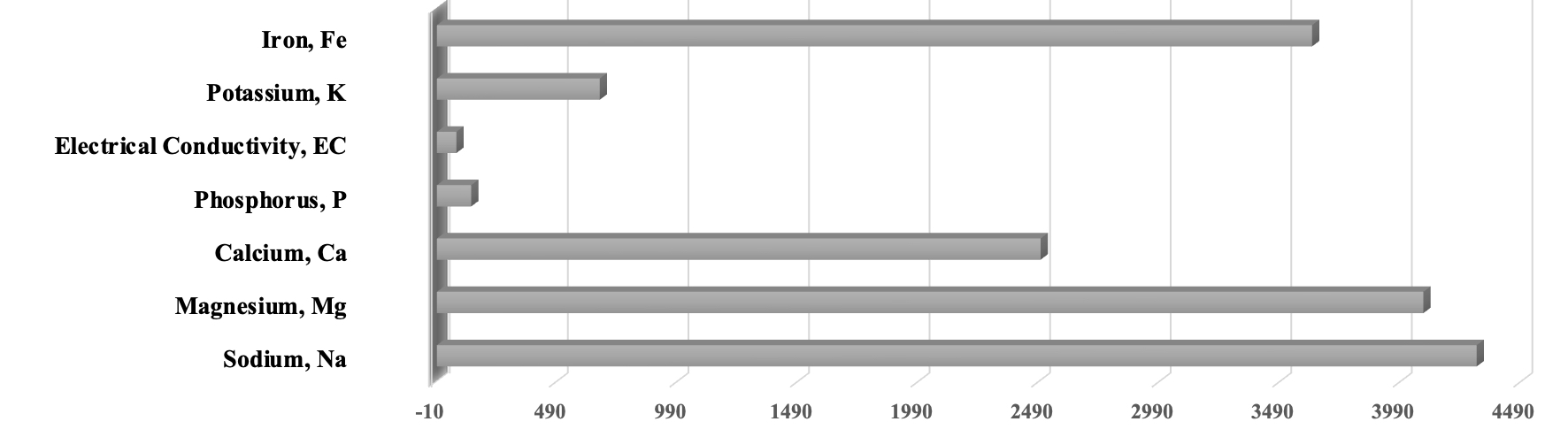}
\caption{The top environmental variables learned from DS combinatorial optimization to reduce the risk of food-borne illness. This figure illustrates the pre-harvest practices and DS generated values that can help to reduce the presence of MDR on poultry farms. Indications from these recommendations can serve as a basis for further analysis and practical experimentation.}
\label{fig::deepSA_mdr}
\end{figure*}

\paragraph{Selection}

To satisfy our objective of finding features that can meet our expectations, we use prediction-based selection at this stage. If our objective is to find combinations that reduce the prediction, we should use $1 - \Lambda$ (prediction probabilities from neural network, $\M$) in the expression. Our objective is achieved by adding a prediction significance threshold, $\omega$ part of prediction to a (1 - $\omega$) part of feature importance, $\Upsilon$. Based on the evaluation of the prediction significance thresholds, $\omega$ between 0.1 and 0.9, 0.6 was found to be the most effective threshold. 

\begin{equation}
\Gamma_i \gets \omega  (1 - \Lambda_i) + (1 - \omega)  \Upsilon_i
\label{eq:relevance}
\end{equation}

Following this, a selection of feature value combinations ($\SN$) is made to provide explanations that meet the highest expectations. Each feature is ranked sequentially, and its score is calculated for every selected value in equation \ref{eq:relevance}. The combinations tree is reduced at every stage with a long-term cut, $\zeta$, which ensures a reduction in computation. Based on experimental evaluations, using $\zeta$ scores of 5 yielded promising results while reducing computation costs.

\begin{equation}
\SN \gets \argmax_{\zeta} \Gamma_{f, v} 
\end{equation}

\section{Data Corpus} \label{dataset}

Our study conducts experiments with two multi-label classification datasets. With the first dataset, our objective is to reduce the multi-drug resistance of three pathogens associated with pre-harvest poultry management practices. Our objective in the second problem is also a multi-label classification, in which the aim is to determine the best combination of post-harvest poultry practices that will ensure the reduction of pathogens.

\subsection{Multi-drug resistance reduction in pre-harvest pastured poultry practices}

This study utilized an existing pastured poultry dataset~\cite{hwang2020farm, pillai2022ensemble}, collected from eleven pastured poultry farms located in the southeastern United States over a four-year period and has been previously described in detail~\cite{ayoola2022preharvest}. The pre-harvest samples (feces and soil) were collected at three different times: immediately following the placement of broilers on pasture, halfway through their pasture stay, and on the day of the flock’s processing. A minimum of 25 grams of sample was collected for each field sample. Three grams of phosphate-buffered saline (PBS) diluted 1:3 with feces and soil were placed within filtered stomacher bags (Seward Laboratories Systems, Inc., West Sussex, UK). The samples were homogenized for 60 seconds, and homogenates were used for downstream culture isolations of \textit{Salmonella}, \textit{Listeria}, and \textit{Campylobacter}. All three pathogens were characterized using the published NARMS protocols and NARMS breakpoints, and isolates that were resistant to three or more antibiotics were considered multi-drug resistant (MDR). The MDR was predicted based on a combination of general poultry management settings (31 farm and management practice variables) and physicochemical variables (Total Carbon, Total Nitrogen, and elemental composition (Al, As, B, Ca, Cd, Cr, Cu, Fe, K, Mg, Mn, Mo, Na, Ni, P, Pb, S, Si, Zn)) as dataset features.

\subsection{Pathogen reduction in post-harvest pastured poultry practices}

In this multi-classification investigation set, features are collected from poultry chicken samples during the processing (Cecal) and postprocessing stages (whole carcass rinse (WCR) immediately following processing, as well as the final product rinse (WCR) after chilling and storage) (Detailed explanation is found in~\cite{rothrock2016antibiotic}]. \textit{Salmonella}, \textit{Listeria}, and \textit{Campylobacter} were identified and analyzed from a number of poultry farm practices, as well as processing, water, freezing, and storage practices from the poultry processing stage, to determine the source of pathogen contamination.

\section{Experiments and Results} \label{experiments}

The two datasets used in this study are highly imbalanced and complex.  The number of negative samples is six times higher than the number of positive samples. For training, we used 90\% of the dataset and for testing, we used 10\% of the dataset. We build our neural networks using TensorFlow (version 2.10)~\cite{tensorflow2015-whitepaper}, Keras (version 2.14.0)~\cite{chollet2015keras}, and PyTorch (version 2.1)~\cite{NEURIPS2019_bdbca288}. It is implemented in Python (version 3.11)~\cite{van1995python} and metrics are calculated using Scikit-Learn (version 1.3) ~\cite{sklearn_api}.

\begin{figure*}[h]
\centering
\includegraphics[width=0.95\textwidth]{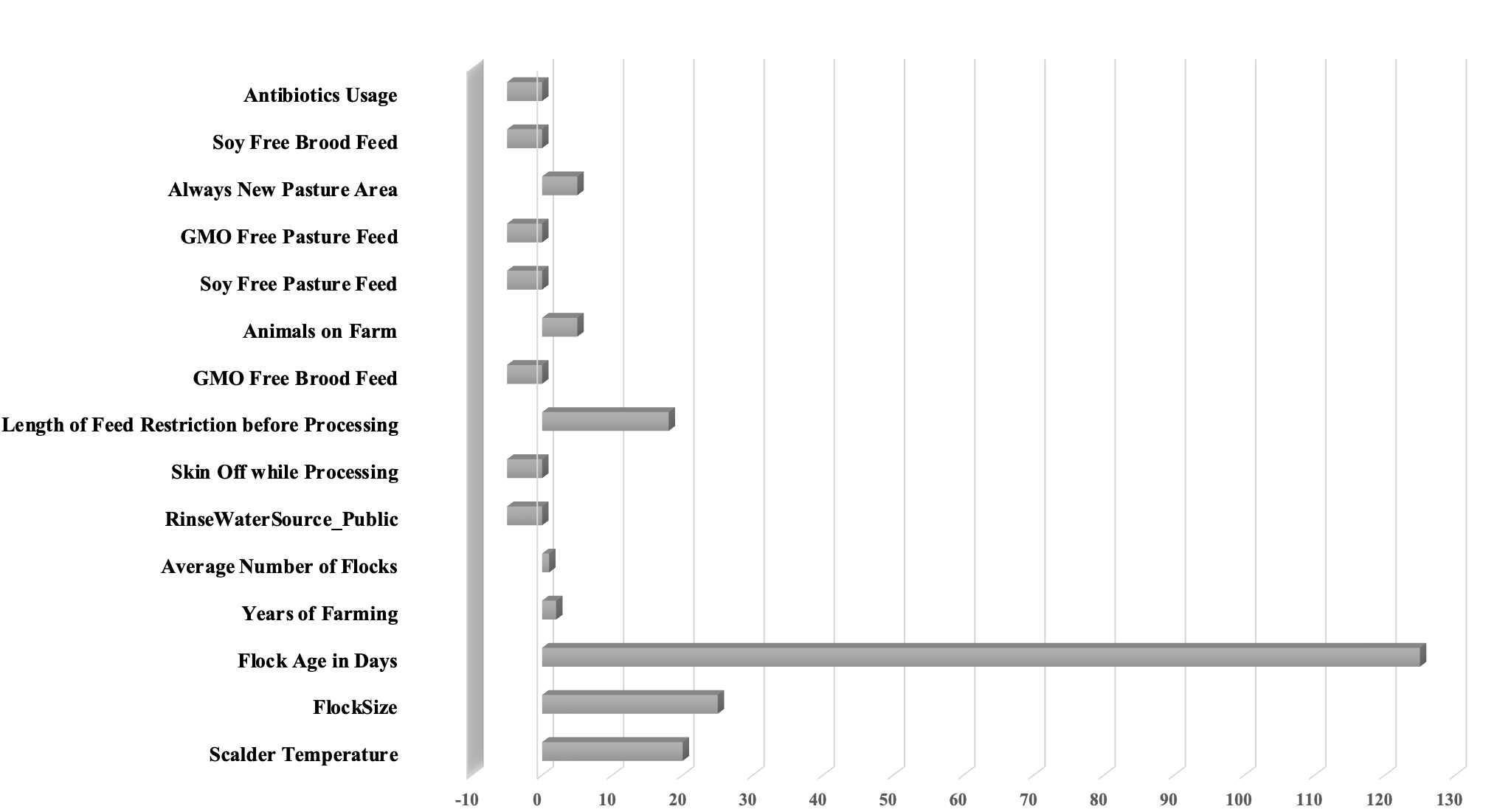}
\caption{The top poultry practices learned from DS combinatorial optimization to reduce the risk of food-borne illness. Figure presents post-harvest practices and DS generated feature values to help reduce pathogen contamination. Certain features produce a negative score, which indicates a reduction in pathogens in the absence of the factors.}
\label{fig::deepSA_pathogen}
\end{figure*}

\paragraph{Baseline} We used a dynamic programming (DP) based optimization design~\cite{Pillai_CO2023} as a baseline to verify the efficiency of our combinatorial optimization model.  Dynamic programming falls under the category of exact algorithms, which guarantee optimal combinations. This algorithm sequentially validate the prediction results for each feature and choose the optimal combination based on them. Dynamic programming has the main disadvantage of being computationally expensive (For $m$ samples and $n$ features, the complexity would be $m^n$.  We have validated both our baseline architecture and our proposed architecture using two multi-label classification problems.

\subsection{Multi-drug Resistance Reduction} 

Figure~\ref{fig:results_mdr} illustrates the pre-harvest poultry farm data set verification results using our proposed DS model and dynamic programming. In this multi-label classification, our objective was to identify farm practices that would reduce multi-drug resistance of \textit{Salmonella}, \textit{Listeria}, and \textit{Campylobacter}. Based on our results, it appears that our proposed model can select best values at early stages for farm variables that will result in a reduction in MDR in all three pathogens. The dynamic programming method searches the entire dataspace in a grid search-based manner, which can produce exact optimal solutions. However, our model can achieve similar results in fewer iterations, thus making it more efficient.

\subsection{Food-Borne Pathogen Reduction}
Our proposed Deep Sensitivity Analysis and Dynamic Programming technique is tested in order to determine the optimal combination of post-harvest poultry farm datasets. In figure~\ref{fig:results_pathogen}, we can see the initial selection of combinations of features and values. We presented experimental results of \textit{Salmonella}, \textit{Listeria}, and \textit{Campylobacter}, as our objective is to identify poultry settings that can significantly reduce the existence of food-borne pathogens. By incorporating variance-based feature importance scores, it is evident that the DeepSensitivity model can reduce pathogen prediction even at the initial stages. In an imbalanced dataset, individual practices that help reduce \textit{Salmonella} were difficult to identify, but was possible with the association of five features The relevance of interactions between features is demonstrated to be one of the most promising aspects of objective-oriented combinatorial optimization through these experiments.

\subsection{Top Significant Agricultural Practices}
 
Figures~\ref{fig::deepSA_mdr} and \ref{fig::deepSA_pathogen} summarize the combinations of feature values that contribute to the reduction of MDR or pathogen prevalence based on agricultural practices and/or setting.

Physical and chemical factors have a profound impact on the prevalence of antimicrobial resistance (AMR) in poultry production~\cite{ayoola2022preharvest}. An effective method for controlling the dissemination of drug resistant strains can be achieved by manipulating these parameters in a thoughtful manner, regardless of the actual patterns of antibiotic use on farms. In our study (figure~\ref{fig::deepSA_mdr}), physico- chemical properties were associated with a high prevalence of MDR. Our findings indicate that increasing sodium levels represents a promising intervention pathway to mitigate AMR in poultry production environments. Elevated sodium concentrations derived from compounds like NaCl or sodium lactate appear strongly inhibitory against bacterial genera with prevalent MDR strains such as \textit{Salmonella}, and \textit{Campylobacter}. Elevated magnesium, iron, and calcium levels across poultry production environments and processing environments could be an effective solution to MDR.

Moving the flock to a new pasture area always reduces AMR similar to the reported reduction of food-borne pathogens. For both brood and pasture stage chickens, pre-harvest data indicate that a GMO- and soy-free feed is the most effective in reducing AMR.

Our analysis (figure~\ref{fig::deepSA_pathogen}) indicates that scalder temperature has the most significant influence on pathogen prevalence during poultry processing. Based on the data, maintaining the scalder temperature at 20 degrees Celsius is  optimal to minimize pathogen loads while meeting production targets. The scalder is a critical equipment that determines the microbiological quality of the final chicken meat. Strict monitoring and control of the scald temperature
is critical to reduce the risk and growth of pathogens.

Another factor identified in this study that influences the presence of pathogens in a farm is the number of birds on the farm. Each grow-out house in a commercial chicken farm typically houses more than 5000 birds. The high number of birds crowded together in an enclosed space greatly aids the transmission of pathogens within a flock through direct contact and fecal-oral transmission routes. Pathogens spread  fast in bigger flocks. Our evaluation  indicates that farms with flock size in the range of 1500 are likely to have low levels of pathogens. Based on  our findings, reducing flock sizes should be an integral part of enhanced pathogen control programs. Pathogen issues in poultry production are exacerbated by large flocks. To tangibly limit infection reservoirs and curtail pathogen transmission vectors within a farm, establishing lower maximum flock sizes based on housing capacity and economics should also be considered in conjunction with optimizing process controls such as scalder temperatures.

Research suggests that older poultry flocks accumulate more pathogens with time, however, our current analysis indicates that longer production cycles may result in lower levels of food-borne pathogens if they are combined with proper farming practices. Specifically, we observed a significant reduction in pathogen levels in flocks harvested around 125 days of age. There are clearly complex interrelationships between flock age, stress levels, immunity, and pathogen exposure that warrant further study. The longer cycle lengths may have contributed to better acclimation and health-centered husbandry. Based on these findings, we must re-examine the role of shortened production timelines in contributing to intensified disease pressures. As well, public water supplies contribute to elevated pathogen exposure. To mitigate contamination risks, a comprehensive root cause analysis covering the entire production ecosystem is essential.

Our study involved pastured poultry chickens that were subjected to various periods of pre-slaughter feed restriction ranging from 8 to 24 hours. Our analysis identifies 18 hours to be the optimal window - sufficiently clearing digestive tract residues before transport and slaughter while avoiding extended starvation stress. Furthermore, we have observed that soy free brood feed, GMO free brood feed, and soy free pasture food are associated with an increase in pathogen prevalence, while GMO free pasture food is associated with a reduction in pathogen prevalence. It is noteworthy that the results of the feed based study differ slightly from those of the pre-harvest study. Moving to a new pasture area is always an effective means of reducing the proliferation of pathogens.

\section{Conclusion} \label{conclusion}

To conclude, our study proposed a deep learning based global sensitivity analysis for the purpose of determining the importance of features in explaining artificial intelligence systems. Also, we demonstrate the applicability of such systems to the problem of combinatorial optimization. To validate the effectiveness of the proposed system, we performed experiments on two multi-label classification problem sets. In addition, our system showed improvements over an optimal dynamic programming-based combinatorial approach. Using such optimization algorithms is vital in agricultural settings to reduce the presence of food-borne pathogens and increase food safety. Explainable AI-based system that can solve combinatorial problems described in this study can be utilized in multiple agricultural and biomedical settings.

\section*{Acknowledgment}
Dataset used in this study is provided by the Agricultural Research Service, USDA CRIS Project ``Reduction of Invasive \textit{Salmonella enterica} in Poultry through Genomics, Phenomics and Field Investigations of Small MultiSpecies Farm Environments'' \#6040-32000-011-00-D. This research was supported by the Agricultural Research Service, USDA NACA project entitled ``Advancing Agricultural Research through High Performance Computing'' \#58-0200-0-002.

\bibliography{references}
\bibliographystyle{IEEEtranS}

\end{document}